\documentclass{INTERSPEECH2023}
\usepackage{amsmath}
\usepackage{pdfpages}
 \interspeechcameraready

\title{Understanding the Impact of Post-Training Quantization on Large Language Models}
\name{Somnath Roy}
\address{
  Freshworks Inc}
\email{somnath.roy@freshworks.com}

\begin{document}

\maketitle
 
\begin{abstract}
Large language models (LLMs) are rapidly increasing in size, with the number of parameters becoming a key factor in the success of many commercial models, such as ChatGPT, Claude, and Bard. Even the recently released publicly accessible models for commercial usage, such as Falcon and Llama2, come equipped with billions of parameters. This significant increase in the number of parameters makes deployment and operation very costly. The remarkable progress in the field of quantization for large neural networks in general and LLMs in particular, has made these models more accessible by enabling them to be deployed on consumer-grade GPUs. Quantized models generally demonstrate comparable performance levels to their unquantized base counterparts. Nonetheless, there exists a notable gap in our comprehensive understanding of how these quantized models respond to hyperparameters, such as temperature, max new tokens, and top\_k, particularly for the next word prediction. 

The present analysis reveals that nf4 and fp4 are equally proficient 4-bit quantization techniques, characterized by similar attributes such as inference speed, memory consumption, and the quality of generated content. Nevertheless, these quantization methods exhibit distinct behaviors at varying temperature settings, both in the context of smaller and larger models. Furthermore, the study identifies nf4 as displaying greater resilience to temperature variations in the case of the llama\_2 series of models at lower temperature, while fp4 and fp4-dq proves to be a more suitable choice for falcon series of models. It is noteworthy that, in general, 4-bit quantized models of varying sizes exhibit higher sensitivity to temperature in the range of 0.5 to 0.8, unlike their unquantized counterparts. Additionally, int8 quantization is associated with significantly slower inference speeds, whereas unquantized bfloat16 models consistently yield the fastest inference speeds across models of all sizes.
\end{abstract}
\noindent\textbf{Index Terms}: post-training quantization, LLM, nf4, fp4, nf4-dq, fp4-dq

\section{Introduction}

With the emergence of the Transformer architecture \cite{vaswani2017attention}, a significant breakthrough was achieved, enabling the effective retention of extensive long-range dependencies in tasks related to natural language processing, speech, and vision. The transformer architecture enables highly parallel training due to sequence parallelism, which makes it possible to pretrain LLMs with hundreds of billions of parameters \cite{brown2020language, chowdhery2022palm, smith2022using}. The Big-bench \cite{srivastava2022beyond} introduced over 200 benchmarks designed to assess the capabilities of Large Language Models(LLMs) through quantification and extrapolation. This diverse and intricately elaborated set of benchmarks significantly contributed to the intensification of the race surrounding LLM development and advancement. 

The widespread adoption of LLMs on a substantial scale gained traction following the successful establishment of ChatGPT (including GPT-3 and subsequent iterations) \cite{brown2020language}. The pre-training of large transformer language models with 7 billion parameters and beyond demands a considerable amount of GPU computation, which can translate to costs amounting to millions of dollars. Such level of expenditure is beyond what academic research and small organizations can typically afford. Despite the high cost of deploying and operating large language models (LLMs), the recent release of the Falcon \cite{penedo2023refinedweb} and Llama2 \cite{touvron2023llama} models has sparked optimism among small organizations and has increased their desire to deploy their own custom LLMs. 

The efficient deployment of decoder only LLMs are challenging in practice because the generative inference proceeds sequentially, where the computation for each token depends on the previously generated tokens \cite{pope2023efficiently}. It is noteworthy that caching the attention key and value tensors of each layer can significantly improve the inference speed of smaller decoder-only models that fit on a single GPU memory. However, this is not possible for models that do not fit into the memory of a single GPU. To address the need for expensive high-end GPUs to support the deployment of these models, diverse forms of quantization have been put forward as potential solutions. The application of quantization methods to transformers emerges as a efficacious approach for mitigating sampling latency, while incurring minimal to negligible impact on overall performance \cite{chen2023accelerating}. Quantization techniques can be mainly characterized into three forms namely - i) quantization aware training \cite{yang2023dynamic, liu2023llm}, ii) quantization aware fine-tuning \cite{dettmers2023qlora, kwon2022alphatuning, dettmers2022llm}, and iii)  post training quantization (PTQ) \cite{frantar2022gptq, yuan2023rptq, lin2023awq}. In \cite{yao2023comprehensive:}, the investigation primarily centers on evaluating the impact of diverse post-training quantization methods, employing perplexity scores as a benchmark. The perplexity scores are computed on datasets such as Wiki \cite{merity2016pointer}, PTB \cite{marcus1993building}, and C4 \cite{raffel2020exploring}, which mostl likely have served as foundational datasets during the training of most of the LLMs. It should be noted that these datasets are predisposed to exhibit favorable perplexity scores across all models, owing to their utilization in model training. Furthermore, it is acknowledged that perplexity, as a metric, may not effectively capture instances of repetitive generation within LLMs. Following outlines the primary contributions of the present study.
\begin{enumerate}
    \item This study offers a systematic examination of the influence exerted by three pivotal hyper-parameters, namely, max new tokens, temperature, and top\_k, on LLMs that have undergone quantization through widely adopted post-training quantization techniques such as \cite{frantar2022gptq}\footnote{https://github.com/IST-DASLab/gptq} (hereafter, gptq) and \cite{dettmers2022llm, dettmers2023qlora}\footnote{https://github.com/TimDettmers/bitsandbytes}\footnote{https://github.com/artidoro/qlora} (hereafter, bitsandbytes).
    \item It explores how these hyper-parameters exert their influence across a range of model sizes, spanning from 3 billion to 70 billion parameters.
    \item The process involves generating a total of 6,300 samples for each quantization method, achieved by constructing ten smaller prompts that encompass a diverse spectrum of domains for every model. 
    \item LLMs typically exhibit a tendency towards repetitive generation, and it is often challenging to discern such repetition through perplexity scores. Therefore, to identify and quantify repetitive generation, the primary metric employed is the number of duplicate content words.
    \item It scrutinizes quantization methods that share similar inference speeds but manifest differing effects on accuracy.
    \item Finally, it aims to discern the optimal quantization method for deployment, considering specific constraints and requirements.
    
\end{enumerate}

\section{Quantization}
Quantization is a well defined mechanism for reducing the number of bits used to represent a value. In the context of large neural network models, quantization reduces the precision of the model's parameters and/or activations. Moreover, it has been found that the quantized large models are often competitive to its base ones in terms of accuracy while reducing the computational requirements. \\
In the context of LLMs, the quantization process can be divided into two types namely i) simulated and, ii) pure quantization. In simulated quantization, some operations are performed in floating-point arithmetic, which requires the dequantization of quantized parameters back to full precision during inference. \cite{shen2020q, zadeh2020gobo, bai2020binarybert,zhang2020ternarybert}. Pure quantization uses integer-only quantization, which eliminates the need for dequantization during inference \cite{kim2021bert,yao2023comprehensive, xiao2023smoothquant, dettmers2023qlora, wu2023understanding}. The main difference between these two process of quantization is shown below in Table 1. 
\begin{table}[h]
\addtolength{\tabcolsep}{-0.02pt}
\begin{center}
\begin{tabular}{|p{2cm}|p{2cm}|p{2cm}|}
 \hline
 \textbf{Features} & \textbf{Simulated Quantization} & \textbf{Pure Quantization} \\
 \hline
 {Operations} & {Floating and Fixed point} & {Fixed-point} \\
 \hline
Need for dequantization & Yes & No \\
\hline
Inference speed & Slower & Comparatively Faster\\
\hline
\end{tabular}
\end{center}
\caption{General understanding of simulated vs. pure quantization in transformer based LLMs}
\end{table}
However, it is crucial to note that pure quantization is a more aggressive approach and can also lead to a greater loss of accuracy. On the other hand, simulated quantization is a conservative approach and can achieve significant speedups without sacrificing too much accuracy. Pure quantization can be further categorized into W8A8 and W4A4, where the weights and activations are quantized to 8-bit integers and 4-bit integers, respectively \cite{wu2023understanding} \cite{yao2023comprehensive}. 

\subsection{GPTQ}
It is a layer-wise quantization method based on the Optimal Brain Quantization (OBQ) \cite{frantar2022optimal}. The goal is to find a quantized weight matrix $\widetilde{W}$ that minimizes the squared error between the quantized layer output $\widetilde{W}X$ and the full-precision layer output $WX$ as shown below. 
\[
\underset{\widetilde{W}}{\operatorname{argmin}}
\lVert{WX}- \widetilde{W}X\rVert^2
\]
The OBQ algorithm iteratively quantizes one weight at a time, while the GPTQ algorithm utilizes a vectorized implementation that allows it to efficiently handle multiple rows of the weight matrix in parallel. This makes GPTQ significantly faster than OBQ, especially for large models.
\subsubsection{GPU Memory Consumption in 4-bit GPTQ Quantization}
It is well-established that the goal of quantization is to deploy LLMs on consumer-grade GPUs having at most 24 GB.The distribution of GPU memory utilised by the models during GPTQ 4-bit quantization is shown below in Table 2. GPTQ quantization has following limitations. 
\begin{itemize}
    \item It is very GPU memory intensive process. 
    \item Even 4-bit quantization of 40B model throws out of memory (OOM) on 80GB A100 GPU machine. Moreover, it is not possible to quantize 7B models on 24GB A10 GPU machines.  
    
\end{itemize}

\begin{table}[h]
\begin{center}
\begin{tabular}{|p{2cm}|p{4cm}|}
 \hline
 \textbf{Model} & \textbf{GPU Memory(GB)}\\
 \hline
 {stablelm\_3b} & {19.54} \\
 \hline
 
 {redpajama\_3b} & {9.58} \\
 \hline
 {falcon\_7b} & {23.64} \\
 \hline

 {llama2\_7b} & {24.83} \\
 \hline
 {llama2\_13b} & {40.46} \\
 \hline

 {falcon\_40b and llama2\_70b} & {OOM on single A100 80GB GPU} \\
 \hline
\end{tabular}
\end{center}
\caption{Distribution of GPU memory consumed by GPTQ 4-bits quantization for different models evaluated on Nvidia A100 80GB GPU machine}
\end{table}

\subsubsection{Layerwise Error induced by GPTQ}
GPTQ 4-bit quantization reduces the size of a model by more than 80\%, i.e., a model of 14 GB is reduced to around 2 GB post quantization. It is important to note that the quantization error introduced by GPTQ is different for different models, as shown in Table 3. This is because the models shown in Table 3 have different architectures, including the number of heads, number of layers, embedding dimension, number of query groups in multi-query attention, block size, and hidden dimension.

\begin{table*}[h]
\centering
\begin{tabular}{|c|c|c|c|c|}
 \hline
 \textbf{Model} & \textbf{mlp.proj} & \textbf{att.proj} & \textbf{mlp.fc} & \textbf{attn.attn}\\
 \hline
 {stablelm\_3b} & {52850.7}&{12638.9}  & {383200.9}  & {844806.3} \\

 {redpajama\_3b} & {23448.1}&{1048.9}  & {137061.9}  & {138947.9}\\
 
 {falcon\_7b} & {19194.83}&{2362.39}  & {149962.4}  & {32886.3}\\

 {llama2\_7b} & {22773.0}&{3198.7}  & {170837.6}  & {248520.0} \\

 {llama2\_13b} & {27829.5}&{5470.5}  & {247389.2}  & {301002.0} \\
 \hline
\end{tabular}
\caption{Quantization error introduced by GPTQ in mlp projection, attention projection, fully connected and attention layers.}
\end{table*}

\subsection{bitsandbytes Quantizations}
bitsandbytes (bnb) provides implementation of five powerful and state-of-the-art quantization techniques namely i) int8, ii) fp4, iii) nf4, iv) fp4-dq\footnote{dq stands for double quantization}, and v) nf4-dq. The int8 quantization procedure\cite{dettmers2022llm} uses vector-wise quantization with separate normalization constants for each inner product in the matrix multiplication. However, they have found around 0.1\% dominant activation outliers that has the potential to degrade the quality especially in bigger LLMs. Therefore, the precision for these dominant outliers are kept in float16. This scheme isolates the outlier feature dimensions into a 16-bit matrix multiplication, while still allowing more than 99.9\% of the values to be multiplied in 8 bits.

QLoRA\cite{dettmers2023qlora} introduced a new data type called 4-bit normal-float (nf4), which is optimal for normally distributed weights, double quantization to reduce the memory footprint, and paged optimization to manage memory spikes. These techniques together yield excellent inference speed without sacrificing the quality of generation. In nf4 quantization, the base model weights are stored in nf4 data type and computation is performed in bfloat16. However, the model weights are dequantized to bfloat16 in the foward pass for inference \cite{bnb2022}. The bnb quantizations compress the model footprint in the range of 40\% (int8) to ~70\% (nf4-dq). It is important to emphasize here that int8 quantization for llama2\_70B throws OOM error on A100 80GB GPU machine. Rest of the details of compressed model size corresponding to bnb quantizations are described in the following sections.  

\section{Experiment}
This section provides a detailed description of the models, prompts, decoding approach, and related hyper-parameters used to generate the data for the analysis.
\subsection{Model Description}
A total of six pre-trained models with 3 billion to 70 billion parameters were selected for next-word prediction. These models are decoder-only, and their architecture-specific details are shown in Table 4. As can be seen, these models differ from each other in terms of the number of heads, number of layers, embedding dimension, number of query groups used in multi-query attention, sequence length, and intermediate size.
\subsection{Prompts Selection and Proposed Hypothesis}
Ten prompts are designed to access the quality and inference speed of pre-trained models for next word generation. These prompts are selected on the simple proposed hypothesis and shown below in Table 5. 

\textbf{Hypothesis 1:} All pre-trained LLMs trained on billions or trillions of tokens can be ideally conceptualized as a large tree, where each node represents a topic and the text continuations associated with that topic. As we traverse down the tree, the text continuations become more specific and focused. Conversely, as we traverse up the tree, the text continuations become more general and abstract. 

\textbf{Hypothesis 2:} The quality of a pre-trained model can be assessed based on its ability to accurately identify the correct topic node and then traverse to the sub-topic node for focused next word prediction.

\begin{table*}[t]
\centering
\begin{tabular}{|c|c|c|c|c|c|c|}
\hline
\textbf{Model} & \textbf{n\_head} & \textbf{n\_layer} & \textbf{embed\_dim} & \textbf{n\_query\_groups\_mqa} & \textbf{block\_size} & \textbf{intermediate\_size}\\
\hline
stablelm\_3b & 32 & 16 & 4096 &32 & 4096 &16384 \\
redpajama\_3b & 32 & 32 & 2560 &32 & 2048 & 10240\\
falcon\_7b & 71 & 32 & 4544 &1 & 2048 & 18176\\
llama2\_7b & 32 & 32 & 4096 &32 & 4096 & 11008\\
llama2\_13b & 40 & 40 & 4096 &32 & 5120 & 13824\\
falcon\_40b & 128 & 60 & 8192 &8 & 2048 & 32768\\
llama2\_70b & 64 & 80 & 8192 &8 & 5120 & 28672\\
\hline

\end{tabular}
\caption{Descrption of most relevant architectural specs of the pre-trained models used during the experiment}
\label{table:example}
\end{table*}

\begin{table*}[t]
\centering
\begin{tabular}{|c|c|}
 \hline
 \textbf{Prompt} & \textbf{General Expected Continuation}\\
 \hline
 {Life in London} & {Travel/Cultural/Work-Related/London specific stuff} \\
 
 {It is easy to be a techie} & {Comparison of techie with other probable roles in tech sector}\\
 
 {Stock brokers are earning} & {Stock brokers and their earning style, sources, etc}\\

 {It looks like written by Shakespeare} & {Shakespeare style text comparison} \\

 {Hello, my name is} & {Chat or Introduction}\\
 {Global warming and AI} & {Global Warming and AI in general as well as their +ive and -ive association}\\
 {Current world order} & {Essay/Discussion/Power and Politics related tow world order}\\
 {Percentage of people adore actors and singers} & {Stats on people following their favourite actors/singers and discussion on the related topic}\\
 {Exercise and eating habits for} & {Eating habits and exercise routine in general (pros and cons) }\\
 {Millennial and genz} & {Comparison and contrast between millennial and genz}\\
 \hline
\end{tabular}
\caption{Prompts Description}
\end{table*}
\subsection{Decoder Description}
The current experiment uses a bare top\_k sampling decoder without any additional features, such as repetition penalty. To assess the model's potential, we use a list of max new tokens, temperature as well as top\_k. The max new tokens, temperature and top\_k are [50, 100, 150, 200, 250, 300, 350, 400, 450, 500], [0.1, 0.2, 0.3, 0.4, 0.5, 0.6, 0.7, 0.8, 1.0] and [1, 5, 10, 20, 50, 100, 200] respectively. The completion text is generated for every quantized models using all the combinations of max new tokens, temperature and top\_k. The reason for high top\_k such as 200 is that it might allow models to choose more diverse, less repetitive, and semantically coherent text.

\section{Analysis}
A total of 6300 (10 prompts $\times$ 10 max new tokens $\times$ 9 temperature $\times$ 7 top\_k) completion text is generated for each quantized model except falcon\_40b and llama2\_70b for 16bit \footnote{Both falcon\_40b and llama2\_70b encounter OOM errors on an 80GB GPU machine. falcon\_40b throws an OOM error when the max new tokens exceeds 400, while llama2\_70b faces an OOM error during the loading process.}. The evaluation of these completion texts is conducted through the computation of counting the duplicate content words, serving as a metric for assessing the quality of the generated text. The content words are the remaining words after removing the stop words. Additionally, the model's size in gigabytes (GB) serves as a key measure for quantifying GPU memory consumption, while tokens/sec is employed as a metric to gauge the model's inference speed.  

\subsection{Memory Consumption and Inference Speed}
\begin{table*}[t]
\centering
\begin{tabular}{|c|c|c|c|c|c|c|}
\hline
\textbf{Model} & \textbf{bnb.nf4} & \textbf{bnb.nf4-dq} & \textbf{bnb.fp4} & \textbf{bnb.f4-dq} & \textbf{bnb.int8} & \textbf{bfloat16}\\
\hline
stablelm\_3b & 3.22 & 3.20 & 3.22 &3.06 &4.68 &7.42 \\
redpajama\_3b & 2.31 & 2.17 & 2.31 &2.17 & 3.52 &  5.60\\
falcon\_7b & 5.72 & 5.37 & 5.72 & 5.37 & 8.71 & 14.50\\
llama2\_7b & 4.58 & 4.27 & 4.58 & 4.27 & 7.82 & 13.53\\
llama2\_13b & 8.83 & 7.8 & 8.83 & 7.8 & 14.2 & 26.23\\
falcon\_40b & 26.40 & 24.64 & 26.55 & 24.64 & 44.52 & 80.85\\
llama2\_70b & 40.23 & 38.2 & 40.4 &38.2 & 70.44 & -\\
\hline

\end{tabular}
\caption{The distribution of memory consumed (lower is better) for all the models for different quantization evaluated on Nvidia A100 80GB GPU machine.}
\label{table:example}
\end{table*}

The utilization of int8 quantization demonstrates a significant reduction in memory consumption, approximately in the range of 40\% to 50\%, when compared to bfloat16, as illustrated in Table 6. Nonetheless, it is important to note that this enhancement is accompanied by a corresponding trade-off in inference speed, with int8 exhibiting a slowdown of roughly 75\% to 80\% in comparison to bfloat16, as indicated in Table 7.

When evaluating memory consumption between the fp4 and nf4 quantization approaches for model sizes up to 13 billion, their distinctions are negligible. However, nf4 quantization exhibits a slight advantage over fp4 in terms of memory consumption for larger models such as falcon\_40b and llama2\_70b. Nevertheless, fp4-dq is found to be better in memory consumption (i.e., takes less memory) across the models compared to its counterpart, as shown in Table 6. It is worth noting that while double quantization offers a clear advantage in memory consumption, it results in an inference speed reduction of approximately 10\% to 25\% compared to the absence of double quantization, as outlined in Table 7.

In conclusion, among the various quantization methods, bfloat16 stands out as the least efficient in terms of memory consumption. However, it excels in terms of inference speed, except in the case of stablelm\_3b.

\begin{table*}[t]
\centering
\begin{tabular}{ccccccc}
\toprule
\textbf{Model} & \textbf{bnb.nf4} & \textbf{bnb.nf4-dq} & \textbf{bnb.fp4} & \textbf{bnb.f4-dq} & \textbf{bnb.int8} & \textbf{bfloat16}\\
\midrule
stablelm\_3b & \textbf{(37.76, 62.79)} & (38.7, 53.37) & \textbf{(42.99, 63.11)} & (38.7, 53.03) & (7.91, 16.81) & (37.76, 49.88) \\
redpajama\_3b & (24.2, 32.29) & (22.59, 27.04) & (25.64, 31.37) & (15.49, 27.08) & (2.52, 3.24) & \textbf{(29.35, 37.85)}\\
falcon\_7b & (29.09, 37.54) & (22.71, 30.04) & (24.77, 37.41) & (22.23, 30.7) & (3.13, 12.63) & \textbf{(35.79, 48.05)}\\
llama2\_7b & (23.09, 29.88) & (19.32, 23.44) & (23.01, 28.65) & (17.85, 23.41) & (1.32, 8.87) & \textbf{(28.39, 36.35)}\\
llama2\_13b & (15.9, 23.14) & (13.22, 18.84) & (12.0, 22.98) & (10.83, 18.22) & (6.49, 7.14) & \textbf{(24.12, 29.34)}\\
falcon\_40b & (11.93, 16.59) & (11.57, 14.51) & (12.12, 16.61) & (10.42, 12.76) & (3.56, 4.63) & \textbf{(12.37, 13.99)}\\
llama2\_70b & (8.67, 10.39) & (6.47, 9.07) & (8.52, 10.23) & (6.39, 8.82) & (2.79, 3.76) & -\\
\bottomrule
\end{tabular}
\caption{The distribution of minimum and maximum inference speed (higher is better) in tokens/sec for different quantization evaluated on Nvidia A100 80GB GPU machine.}
\label{table:example}
\end{table*}

\subsection{Temperature vs. Quality of Generation}
A common pattern emerges within all quantization approaches, wherein an increase in temperature correlates with an elevation in number of duplicate content words except for bfloat16. However, it is worth noting that some models are more sensitive to even temperature lower than 0.5 compared to others. 

When comparing the performance of stablelm\_3b and redpajama\_3b models, it becomes evident that the fp4 and nf4-dq quantization methods exhibit suboptimal results, characterized by an increased occurrence of duplicate words at lower temperature settings. However, the situation varies when considering falcon models, where nf4 quantization consistently demonstrates inferior performance across the entire temperature spectrum in comparison to other quantization methods.

In contrast, when assessing llama2 models, the situation becomes more nuanced, with most quantization approaches contributing significantly to repetitive generation. In this context, determining a clear front-runner among these methods proves to be a challenging task. Nevertheless, it is noteworthy that for the llama2\_70b model, both fp4 and fp4-dq quantization methods outshine the others in terms of performance.

The analysis reveals that the int8-quantized model demonstrates effective control over the occurrence of duplicate content words for both llama2\_13b and llama2\_70b, effectively limiting them in the range of 40. In contrast, the bfloat16 models exhibit a characteristic of independence from temperature scaling, as they consistently generate a comparable number of repetitive words across all temperature settings except redpajama\_3b.

\subsection{Max Returned Tokens vs. Quality of Generation}
The term max returned tokens encompasses the combined value of max new tokens and the length of the input prompt in terms of tokens. the analyis reveals that an the count of duplicate words generated linearly increases with the increase of max returned tokens across all models and quantization methods. 

\subsection{Top\_k vs. Quality of Generation}
The analysis offers a somewhat surprising insight, indicating that setting top\_k equal to 1 tends to result in the lowest occurrence of duplicate words across models and quantization methods. Nonetheless, it's noteworthy that this effect reaches a point of saturation and loses distinctiveness when top\_k is equal to or greater than 5.

\subsection{Overall Comparison}
In terms of the average number of duplicate content words\footnote{The total number of content words generated for the unquantized model lies in the range of 1.34M to 1.45M and the maximum duplicate number of words is around 80K.}
generated in absolute terms, our analysis reveals the following insights:
\begin{itemize}
    \item For fp4 and fp4-dq compared to nf4 and nf4-dq across various models (except llama2 series), there is a consistent reduction in repetitive generation, typically ranging from 12\% to 20\% relative.
    \item In the case of nf4 and nf4-dq for llama2 models of different sizes, there is a more noticeable advantage, with relative reduction of 9\% to 11\% in repetitive generation. 
    \item Int8 quantization has a more pronounced limitation on the number of generated words, producing approximately 30-50\% fewer content words than 4-bit quantization. Additionally, it produces 25-40\% more duplicate content words relative to 4-bit quantization at normalized scale.
    \item When comparing bfloat16 with 4-bit quantization, it's noteworthy that bfloat16 generally produces more number of content words, often ranging from approximately 3\% to 10\%. Nonetheless, bfloat16 tends to generate a marginally higher duplicate words, indicating relative inferiority of 1\% to 3.5\% with 4bit quantization.  
    
\end{itemize}
The computation of average perplexity scores, with a token stride of 512, is conducted for all quantization levels across each model. An examination of these scores reveals that the perplexity values for all models reside within a relatively constrained range, typically ranging from 12 to 15. Consequently, it is discerned that perplexity, within this context, may not serve as a suitable metric for assessing the quality of the generated text.

\section{Conclusions}
In scenarios where GPU memory is not a limiting factor and the utmost priority is placed on achieving both high inference speed and accuracy, it is advisable to prioritize the utilization of bfloat16 for models up to 7 billions. This preference arises due to its reduced susceptibility to variations in temperature and max new tokens. Moreover, model upto 7 billion size effectively fits into a consumer grade GPU machine. Alternatively, nf4 and fp4 serves as the default choice for individuals seeking a balance between GPU utilization, accuracy and inference speed, thus offering a middle-ground solution that combines all aspects effectively.

It's worth noting that the adoption of double quantization, such as fp4-dq and nf4-dq, can lead to a marginal reduction in memory footprint. However, it is accompanied by a relatively decreased inference speed. Hence, the recommendation leans toward using quantization without the doubling approach. Additionally, when considering the nf4 and fp4 precision combination, it is recommended to use a temperature of less than 0.5, exactly 1.0, or a combination of these values to achieve optimal performance.

The current evaluation does not consider int8 to be a feasible alternative to other quantization methods. While int8 reduces memory usage, it significantly slows down inference and produces around 30-50\% fewer words than other quantization methods.

It is important to note that the current experiment did not achieve satisfactory results in terms of accuracy and inference speed when using gptq 4-bit quantization. Further investigation is needed to replicate the comparable performance that has been reported in other studies\footnote{https://github.com/PanQiWei/AutoGPTQ}. Therefore, this result is not included in the analysis presented.

\section{Limitations and Future Work}
The current study is conducted on 7 models ranging in size from 3 billion to 70 billion parameters, and 10 prompts are used for next-word prediction using various combinations of hyperparameters. Further study with more models (containing $\leq$ 1 billion parameters) and prompts might provide more insights into the effects of these hyperparameters on relatively smaller quantized LLMs.

Future work will focus on the primary causes of repetitive generation and their relationship to Hypothesis 1 and Hypothesis 2. Moreover, the results show that falcon has a faster inference speed than llama2 in the 7B category. However, falcon has a higher number of overall parameters than llama2. Therefore, future research will focus on model-specific factors that affect inference speed.

\bibliographystyle{IEEEtran}
\bibliography{mybib}

\begin{thebibliography}{10}
\providecommand{\url}[1]{#1}
\csname url@samestyle\endcsname
\providecommand{\newblock}{\relax}
\providecommand{\bibinfo}[2]{#2}
\providecommand{\BIBentrySTDinterwordspacing}{\spaceskip=0pt\relax}
\providecommand{\BIBentryALTinterwordstretchfactor}{4}
\providecommand{\BIBentryALTinterwordspacing}{\spaceskip=\fontdimen2\font plus
\BIBentryALTinterwordstretchfactor\fontdimen3\font minus
  \fontdimen4\font\relax}
\providecommand{\BIBforeignlanguage}[2]{{%
\expandafter\ifx\csname l@#1\endcsname\relax
\typeout{** WARNING: IEEEtran.bst: No hyphenation pattern has been}%
\typeout{** loaded for the language `#1'. Using the pattern for}%
\typeout{** the default language instead.}%
\else
\language=\csname l@#1\endcsname
\fi
#2}}
\providecommand{\BIBdecl}{\relax}
\BIBdecl

\bibitem{vaswani2017attention}
A.~Vaswani, N.~Shazeer, N.~Parmar, J.~Uszkoreit, L.~Jones, A.~N. Gomez,
  {\L}.~Kaiser, and I.~Polosukhin, ``Attention is all you need,''
  \emph{Advances in neural information processing systems}, vol.~30, 2017.

\bibitem{brown2020language}
T.~Brown, B.~Mann, N.~Ryder, M.~Subbiah, J.~D. Kaplan, P.~Dhariwal,
  A.~Neelakantan, P.~Shyam, G.~Sastry, A.~Askell \emph{et~al.}, ``Language
  models are few-shot learners,'' \emph{Advances in neural information
  processing systems}, vol.~33, pp. 1877--1901, 2020.

\bibitem{chowdhery2022palm}
A.~Chowdhery, S.~Narang, J.~Devlin, M.~Bosma, G.~Mishra, A.~Roberts, P.~Barham,
  H.~W. Chung, C.~Sutton, S.~Gehrmann \emph{et~al.}, ``Palm: Scaling language
  modeling with pathways,'' \emph{arXiv preprint arXiv:2204.02311}, 2022.

\bibitem{smith2022using}
S.~Smith, M.~Patwary, B.~Norick, P.~LeGresley, S.~Rajbhandari, J.~Casper,
  Z.~Liu, S.~Prabhumoye, G.~Zerveas, V.~Korthikanti \emph{et~al.}, ``Using
  deepspeed and megatron to train megatron-turing nlg 530b, a large-scale
  generative language model,'' \emph{arXiv preprint arXiv:2201.11990}, 2022.

\bibitem{srivastava2022beyond}
A.~Srivastava, A.~Rastogi, A.~Rao, A.~A.~M. Shoeb, A.~Abid, A.~Fisch, A.~R.
  Brown, A.~Santoro, A.~Gupta, A.~Garriga-Alonso \emph{et~al.}, ``Beyond the
  imitation game: Quantifying and extrapolating the capabilities of language
  models,'' \emph{arXiv preprint arXiv:2206.04615}, 2022.

\bibitem{penedo2023refinedweb}
G.~Penedo, Q.~Malartic, D.~Hesslow, R.~Cojocaru, A.~Cappelli, H.~Alobeidli,
  B.~Pannier, E.~Almazrouei, and J.~Launay, ``The refinedweb dataset for falcon
  llm: outperforming curated corpora with web data, and web data only,''
  \emph{arXiv preprint arXiv:2306.01116}, 2023.

\bibitem{touvron2023llama}
H.~Touvron, L.~Martin, K.~Stone, P.~Albert, A.~Almahairi, Y.~Babaei,
  N.~Bashlykov, S.~Batra, P.~Bhargava, S.~Bhosale \emph{et~al.}, ``Llama 2:
  Open foundation and fine-tuned chat models,'' \emph{arXiv preprint
  arXiv:2307.09288}, 2023.

\bibitem{pope2023efficiently}
R.~Pope, S.~Douglas, A.~Chowdhery, J.~Devlin, J.~Bradbury, J.~Heek, K.~Xiao,
  S.~Agrawal, and J.~Dean, ``Efficiently scaling transformer inference,''
  \emph{Proceedings of Machine Learning and Systems}, vol.~5, 2023.

\bibitem{chen2023accelerating}
C.~Chen, S.~Borgeaud, G.~Irving, J.-B. Lespiau, L.~Sifre, and J.~Jumper,
  ``Accelerating large language model decoding with speculative sampling,''
  \emph{arXiv preprint arXiv:2302.01318}, 2023.

\bibitem{yang2023dynamic}
G.~Yang, D.~Lo, R.~Mullins, and Y.~Zhao, ``Dynamic stashing quantization for
  efficient transformer training,'' \emph{arXiv preprint arXiv:2303.05295},
  2023.

\bibitem{liu2023llm}
Z.~Liu, B.~Oguz, C.~Zhao, E.~Chang, P.~Stock, Y.~Mehdad, Y.~Shi,
  R.~Krishnamoorthi, and V.~Chandra, ``Llm-qat: Data-free quantization aware
  training for large language models,'' \emph{arXiv preprint arXiv:2305.17888},
  2023.

\bibitem{dettmers2023qlora}
T.~Dettmers, A.~Pagnoni, A.~Holtzman, and L.~Zettlemoyer, ``Qlora: Efficient
  finetuning of quantized llms,'' \emph{arXiv preprint arXiv:2305.14314}, 2023.

\bibitem{kwon2022alphatuning}
S.~J. Kwon, J.~Kim, J.~Bae, K.~M. Yoo, J.-H. Kim, B.~Park, B.~Kim, J.-W. Ha,
  N.~Sung, and D.~Lee, ``Alphatuning: Quantization-aware parameter-efficient
  adaptation of large-scale pre-trained language models,'' \emph{arXiv preprint
  arXiv:2210.03858}, 2022.

\bibitem{dettmers2022llm}
T.~Dettmers, M.~Lewis, Y.~Belkada, and L.~Zettlemoyer, ``Llm. int8 (): 8-bit
  matrix multiplication for transformers at scale,'' \emph{arXiv preprint
  arXiv:2208.07339}, 2022.

\bibitem{frantar2022gptq}
E.~Frantar, S.~Ashkboos, T.~Hoefler, and D.~Alistarh, ``Gptq: Accurate
  post-training quantization for generative pre-trained transformers,''
  \emph{arXiv preprint arXiv:2210.17323}, 2022.

\bibitem{yuan2023rptq}
Z.~Yuan, L.~Niu, J.~Liu, W.~Liu, X.~Wang, Y.~Shang, G.~Sun, Q.~Wu, J.~Wu, and
  B.~Wu, ``Rptq: Reorder-based post-training quantization for large language
  models,'' \emph{arXiv preprint arXiv:2304.01089}, 2023.

\bibitem{lin2023awq}
J.~Lin, J.~Tang, H.~Tang, S.~Yang, X.~Dang, and S.~Han, ``Awq: Activation-aware
  weight quantization for llm compression and acceleration,'' \emph{arXiv
  preprint arXiv:2306.00978}, 2023.

\bibitem{yao2023comprehensive:}
Z.~Yao, C.~Li, X.~Wu, S.~Youn, and Y.~He, ``A comprehensive study on
  post-training quantization for large language models,'' \emph{arXiv preprint
  arXiv:2303.08302}, 2023.

\bibitem{merity2016pointer}
S.~Merity, C.~Xiong, J.~Bradbury, and R.~Socher, ``Pointer sentinel mixture
  models,'' \emph{arXiv preprint arXiv:1609.07843}, 2016.

\bibitem{marcus1993building}
M.~Marcus, B.~Santorini, and M.~A. Marcinkiewicz, ``Building a large annotated
  corpus of english: The penn treebank,'' 1993.

\bibitem{raffel2020exploring}
C.~Raffel, N.~Shazeer, A.~Roberts, K.~Lee, S.~Narang, M.~Matena, Y.~Zhou,
  W.~Li, and P.~J. Liu, ``Exploring the limits of transfer learning with a
  unified text-to-text transformer,'' \emph{The Journal of Machine Learning
  Research}, vol.~21, no.~1, pp. 5485--5551, 2020.

\bibitem{shen2020q}
S.~Shen, Z.~Dong, J.~Ye, L.~Ma, Z.~Yao, A.~Gholami, M.~W. Mahoney, and
  K.~Keutzer, ``Q-bert: Hessian based ultra low precision quantization of
  bert,'' in \emph{Proceedings of the AAAI Conference on Artificial
  Intelligence}, vol.~34, no.~05, 2020, pp. 8815--8821.

\bibitem{zadeh2020gobo}
A.~H. Zadeh, I.~Edo, O.~M. Awad, and A.~Moshovos, ``Gobo: Quantizing
  attention-based nlp models for low latency and energy efficient inference,''
  in \emph{2020 53rd Annual IEEE/ACM International Symposium on
  Microarchitecture (MICRO)}.\hskip 1em plus 0.5em minus 0.4em\relax IEEE,
  2020, pp. 811--824.

\bibitem{bai2020binarybert}
H.~Bai, W.~Zhang, L.~Hou, L.~Shang, J.~Jin, X.~Jiang, Q.~Liu, M.~Lyu, and
  I.~King, ``Binarybert: Pushing the limit of bert quantization,'' \emph{arXiv
  preprint arXiv:2012.15701}, 2020.

\bibitem{zhang2020ternarybert}
W.~Zhang, L.~Hou, Y.~Yin, L.~Shang, X.~Chen, X.~Jiang, and Q.~Liu,
  ``Ternarybert: Distillation-aware ultra-low bit bert,'' \emph{arXiv preprint
  arXiv:2009.12812}, 2020.

\bibitem{kim2021bert}
S.~Kim, A.~Gholami, Z.~Yao, M.~W. Mahoney, and K.~Keutzer, ``I-bert:
  Integer-only bert quantization,'' in \emph{International conference on
  machine learning}.\hskip 1em plus 0.5em minus 0.4em\relax PMLR, 2021, pp.
  5506--5518.

\bibitem{yao2023comprehensive}
Z.~Yao, C.~Li, X.~Wu, S.~Youn, and Y.~He, ``A comprehensive study on
  post-training quantization for large language models,'' \emph{arXiv preprint
  arXiv:2303.08302}, 2023.

\bibitem{xiao2023smoothquant}
G.~Xiao, J.~Lin, M.~Seznec, H.~Wu, J.~Demouth, and S.~Han, ``Smoothquant:
  Accurate and efficient post-training quantization for large language
  models,'' in \emph{International Conference on Machine Learning}.\hskip 1em
  plus 0.5em minus 0.4em\relax PMLR, 2023, pp. 38\,087--38\,099.

\bibitem{wu2023understanding}
X.~Wu, C.~Li, R.~Y. Aminabadi, Z.~Yao, and Y.~He, ``Understanding int4
  quantization for transformer models: Latency speedup, composability, and
  failure cases,'' \emph{arXiv preprint arXiv:2301.12017}, 2023.

\bibitem{frantar2022optimal}
E.~Frantar and D.~Alistarh, ``Optimal brain compression: A framework for
  accurate post-training quantization and pruning,'' \emph{Advances in Neural
  Information Processing Systems}, vol.~35, pp. 4475--4488, 2022.

\bibitem{bnb2022}
\BIBentryALTinterwordspacing
Y.~Belkada, T.~Dettmers, A.~Pagnoni, S.~Gugger, and S.~Mangrulkar, ``Making
  llms even more accessible with bitsandbytes, 4-bit quantization and qlora,''
  2023. [Online]. Available:
  \url{https://huggingface.co/blog/4bit-transformers-bitsandbytes}
\BIBentrySTDinterwordspacing

\end{thebibliography}

% \appendix
% \section{Appendices}
% \clearpage
% \includepdf[pages={1-3}]{temp4bits.pdf}
% \includepdf[pages={1-3}]{mrt4bits.pdf}
% \includepdf[pages={1-3}]{topk4bits.pdf}

% \includepdf[pages={1-2}]{tempint8.pdf}
% \includepdf[pages={1-2}]{mrtint8.pdf}
% \includepdf[pages={1-2}]{topkint8.pdf}

% \includepdf[pages={1-2}]{tempbfloat16.pdf}
% \includepdf[pages={1-2}]{mrtbfloat16.pdf}
% \includepdf[pages={1-2}]{topkbfloat16.pdf}

\end{document}